\documentclass[10pt,twocolumn,letterpaper]{article}

\usepackage{iccv}
\usepackage{times}
\usepackage{epsfig}
\usepackage{graphicx}
\usepackage{amsmath}
\usepackage{amssymb}
\usepackage{booktabs}
\usepackage{soul}
\usepackage{color}
\usepackage[accsupp]{axessibility} 

\usepackage[pagebackref=true,breaklinks=true,letterpaper=true,colorlinks,bookmarks=false]{hyperref}
\usepackage[capitalize]{cleveref}
\crefname{section}{Sec.}{Secs.}
\Crefname{section}{Section}{Sections}
\Crefname{table}{Table}{Tables}
\crefname{table}{Tab.}{Tabs.}
\newcommand{\first}[1]{\textbf{#1}}

\newcommand\blfootnote[1]{%
  \begingroup
  \renewcommand\thefootnote{}\footnote{#1}%
  \addtocounter{footnote}{-1}%
  \endgroup
}

\iccvfinalcopy 

\def\iccvPaperID{****} 

\ificcvfinal\fi

\begin{document}

\title{Holistic Geometric Feature Learning for Structured Reconstruction}


\author{
Ziqiong Lu$^{*}$ \quad \quad Linxi Huan$^{*}$ \quad \quad Qiyuan Ma \quad \quad Xianwei Zheng$^{\dagger}$ \\
The State Key Lab. LIESMARS, Wuhan University \\
{\tt\small \{zql2018, whu\_hlx, qiyuanma, zhengxw\}@whu.edu.cn}
}

\maketitle
\blfootnote{
$^*$Equal contribution.}
\blfootnote{
$^{\dagger}$Corresponding author.}
\ificcvfinal\thispagestyle{empty}\fi

	\begin{abstract}

		The inference of topological principles is a key problem in structured reconstruction. We observe that wrongly predicted topological relationships are often incurred by the lack of holistic geometry clues in low-level features. Inspired by the fact that massive signals can be compactly described with frequency analysis, we experimentally explore the efficiency and tendency of learning structure geometry in the frequency domain. Accordingly, we propose a frequency-domain feature learning strategy (F-Learn) to fuse scattered geometric fragments holistically for topology-intact structure reasoning. Benefiting from the parsimonious design, the F-Learn strategy can be easily deployed into a deep reconstructor with a lightweight model modification. Experiments demonstrate that the F-Learn strategy can effectively introduce structure awareness into geometric primitive detection and topology inference, bringing significant performance improvement to final structured reconstruction. Code and pre-trained models are available at \href{https://github.com/Geo-Tell/F-Learn}{https://github.com/Geo-Tell/F-Learn}.
	\end{abstract}
 
	\section{Introduction}
	\label{sec:intro}
	The structured reconstruction models the shape grammar/topology that depicts procedural shape generation~\cite{nauata2020vectorizing}, which can facilitate various downstream vision tasks, such as feature matching~\cite{li2021locally, zheng2022smoothly}, 3D modeling~\cite{nash2020polygen, wu2021deepcad} and 3D scene understanding~\cite{zou2018layoutnet, zhang2020conv}. The recovery of a shape topology is generally realized via two steps: (i) geometric primitive detection and (ii) graph inference based on extracted geometric primitives (e.g., corners and edges). Following this primitive-to-structure principle, previous researchers usually resort to heatmaps of corners and edges for subsequent structure reasoning~\cite{lee2017roomnet, zou2018layoutnet, zhang2020conv, chen2022heat}. With low-level parsing tasks promoted by deep learning techniques, many works have greatly improved the performance of high-level structure recovery in recent years~\cite{zhang2020conv,nauata2020vectorizing, chen2022heat}.
	
\begin{figure*}[t]
		\centering
		\includegraphics[width=\linewidth]{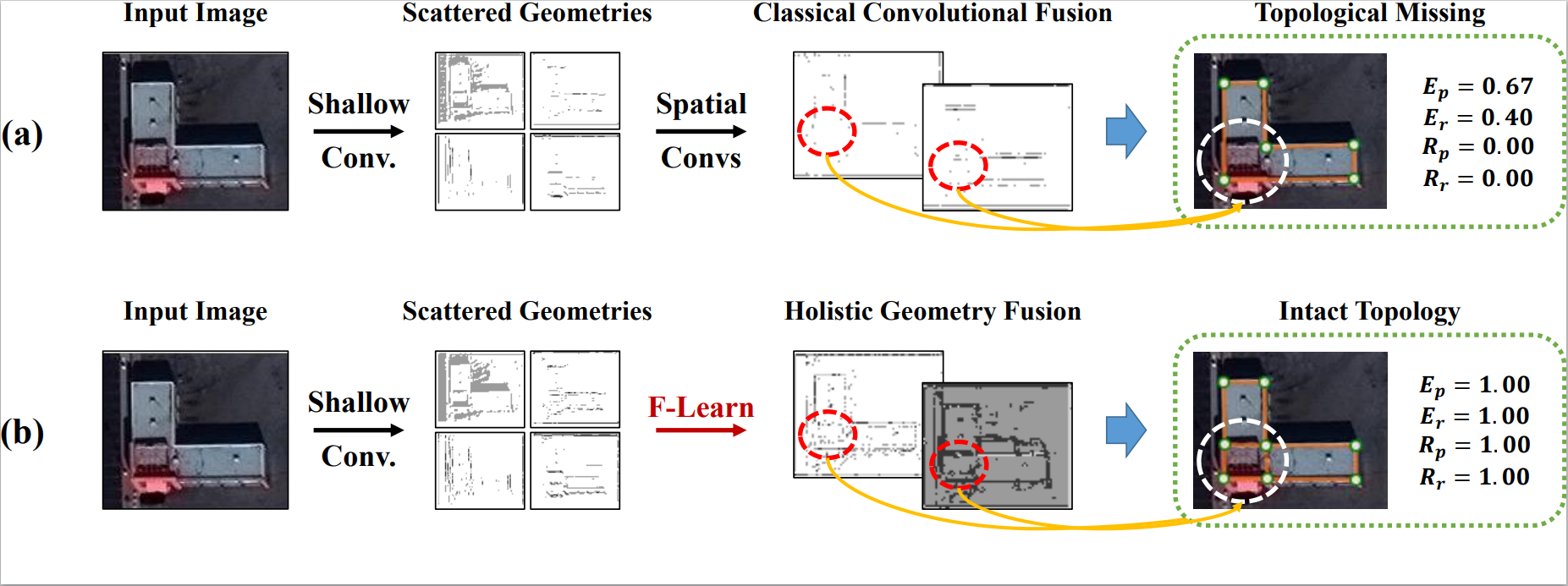}
		\caption{Comparison of learning low-level features in the space and frequency domains. (a) Structure reconstruction with low-level geometric features learned \textbf{in the space domain}. The inefficient fusion of low-level geometric fragments loses the structure clues of the bottom left roof region and consequently results in missing topological relations in the final reconstructed structure graph. (b) Structure reconstruction with low-level geometric features learned \textbf{with the proposed frequency-domain feature learning strategy} (F-Learn). With the geometric features compactly processed in the frequency domain, our F-Learn strategy effectively achieves holistic geometry fusion for inferring the right topological principle for structure reconstruction. $E_p$ and $E_r$ denote the scores of precision and recall for detected edges, while $R_p$ and $R_r$ refer to those of reconstructed regions.}
		\label{fig:ins_info_learn}
	\end{figure*}

	Researchers commonly dedicate to graph inference with geometric features generated by proven hierarchical backbones (e.g., ResNet~\cite{he2016deep}) for structured reconstruction. However, incorrect topology inference remains a significant problem due to the lack of holistic geometric clues in the low-level features. Specifically, the geometric fragments extracted by shallow layers can hardly be fused into structurally informative features, leading to wrongly reasoned topological principles (as shown in~\cref{fig:ins_info_learn} (a)). Different from the prior arts that focus on graph inference, this paper turns to study an efficient strategy for holistically learning structure-related features with low-level geometric fragments.
	
	Low-level geometric features fundamentally support high-level feature extraction and structure recovery. Specifically, the low-level shallow-layer feature maps provide information for precise geometry localization, which is essential for high-level primitive detection and topology inference. Restricted by the limited receptive field, it is hard to holistically capture structure geometries in the shallow layers. We consider that the low efficiency of processing low-level geometric fragments in the space domain is the root cause.
	
	Inspired by the fact that low-level features can be compactly encoded in the frequency domain, we are interested in achieving efficient geometry feature learning with frequency analysis. In the frequency map of a given feature map, every value encodes the global information of the feature map regarding the corresponding frequency. Therefore, performing convolution operations in the frequency domain actually means directly combining the spatial information in a holistic manner. 
 
    To be concrete, given a set of low-level feature maps $\{f_n(x,y)\}$ and the corresponding frequency map $\{F_n(u,v)\}$, the information encoded at $(u_i,v_j)$ describes a certain changing pattern of signals in the space domain. When applying a convolution operation to a frequency position $(u_i,v_j)$, the feature components of related frequencies in every map of $\{f_n(x,y)\}$ will be merged across the channel dimension in the space domain. As the low-level geometric information generally belongs to high-frequency signals, a frequency-domain convolution directly realizes the integration of geometric primitives scatted in $\{f_n(x,y)\}$. The frequency-domain convolution thereby works more efficiently in holistic geometry learning than the space-domain counterpart, which combines local geometric clues without a global view.
	
	Based on the discussion above, we propose a frequency-domain feature learning strategy (F-Learn) to efficiently extract holistic geometry features for guiding the inference of structure topology. Working in the frequency domain, the F-Learn strategy holistically fuses the separated geometric primitives into structurally informative features. The proposed F-Learn strategy can be readily applied to a primitive-to-structure framework for structured reconstruction.
	
	In summary, the main contributions of this paper are:
	\begin{enumerate}
		\item[-] Exploration with a simple geometry recovery task sheds light on the difference in learning tendency between frequency- and space-domain convolutions. The results also validate the high efficiency of frequency-domain convolution in learning holistic geometry. 
		
		\item[-] A parsimonious frequency-domain feature learning strategy (F-Learn) is proposed to generate structurally informative geometry features for structured reconstruction.
		
		\item[-] Experiments on vectorizing world buildings demonstrate that our F-Learn strategy greatly improves the performance of structured reconstruction.
	\end{enumerate}

	\section{Related Work}
	\label{part:related-work}
	\subsection{Geometric Primitive Detection}
	The detection of geometric primitives is a long-standing vision task that has been extensively explored with hand-crafted descriptors in the early stage~\cite{roberts1963machine, gonzales1987digital}. Recently, with the learning ability of hierarchical features, deep learning models have significantly promoted the completeness and precision of detected geometric primitives. A popular way for low-level geometry extraction is to perform heatmap regression or pixel-wise binary classification. The heatmap regression technique is widely adopted in corner detection. For instance, methods like CornerNet~\cite{law2018cornernet} and CenterNet~\cite{duan2019centernet} learn corner heatmaps for the downstream task of object detection. Binary classification is often used to localize edge pixels. Representative edge detectors like HED~\cite{xie2017hed} and RCF~\cite{liu2019richer} focus on fusing multi-level features for edge pixel classification, while some works pay attention to learning crisp edges or semantic contours~\cite{Yang2016Object, deng2018learning, huan2021unmixing}. 
 
 Compared to the pixel-wise inference of corners and edges, the detection of line primitives is more structural as a line segment should be defined by two endpoints~\cite{duda1972use, kamat1998complete}. Since the task of wireframe parsing was introduced by~\cite{huang2018learning}, increasing interest has been witnessed in inferring line segment candidates with learned junctions~\cite{lcnn, xue2019learning, lin2020deep}. With the advance in low- and mid-level primitive detection, researchers have started to make efforts on structured reconstruction in recent years. 
	\subsection{Structured Reconstruction}
	The structured reconstruction requires reasoning the overall topology of a given shape. Generally, structured reconstruction is a large research field that contains various tasks ranging from 3D object CAD modeling~\cite{wu2021deepcad}, layout estimation~\cite{lee2017roomnet, zou2018layoutnet}, to roof extraction~\cite{zhang2021structured}. For instance-level structure reconstruction, many methods resort to generative models~\cite{ wu2021deepcad} or detecting key points with fixed topologies (e.g., skeletons of human bodies and hands)~\cite{xiao2018simple, zimmermann2017learning}. 
 
 As for scene-level structural modeling, the reconstruction targets usually belong to special semantic-related structures, such as planar building roofs and indoor floorplans. For these tasks, the derived structures have to match semantic regions with robustness to other irrelevant structure information. For example, room layout estimation has to ensure that connected lines can form wall regions without extra lines from windows and doors. Related approaches can be roughly categorized into two kinds: (i) the local primitive based and (ii) the global information based. Pioneering local primitive based works derive structural representation from an image by post-processing heatmaps of corners and edges~\cite{lee2017roomnet, zou2018layoutnet}.  
 
 With respect to the global information based methods, it is a popular choice to form the final structure graph with structure-related regions segmented from the input image~\cite{zeng2019deep, stekovic2021montefloor}. Different from the segmentation based methods, recent studies consider a holistic graph inference with geometric primitives as nodes for high-level planer extraction. For example, Conv-MPN~\cite{zhang2020conv} adopts a variant of the graph neural network to pass messages across the whole graph, and HEAT~\cite{chen2022heat} learns the topological pattern of edges with an attention transformer. Beyond exploring the high-level information in structure recovery, we are interested in improving the efficiency of exploiting low-level geometric features, which serve as the foundation of accurate structure topology inference.
	
	\subsection{Frequency Analysis in Deep Learning}
	Frequency analysis is a technique that provides a compressed representation of signals in the frequency domain. Not only commonly used in classical digital signal processing, but frequency analysis also remains powerful in the era of deep learning.~\cite{xu2019frequency} and~\cite{basri2020frequency} investigate the training and generalization of neural networks via frequency analysis.~\cite{xu2020learning} analyzes the spectral bias of deep models regarding multiple vision parsing tasks. ~\cite{dzanic2020fourier} applies frequency analysis to generated images for improving image synthesis quality, while~\cite{jiang2021focal} introduces a focal frequency loss that forces generative models to learn hard frequencies.~\cite{qin2021fcanet} designs a frequency channel attention mechanism to compress channel-wise information with scalars.~\cite{chen2016compressing} and~\cite{wang2018packing} utilize frequency analysis in model compression. More applications of the frequency-domain representation can also be found in domain adaptation~\cite{huang2021fsdr, huang2021rda} and position embedding~\cite{misra2021end, tancik2020fourier}. 
    
    Although frequency analysis has been combined with deep learning techniques in recent years, it is generally used as a mature tool in prior arts without a deep investigation of the reasons why frequency analysis is useful. To this end, we experimentally studied the behavior tendency of frequency-domain convolutions in learning holistic geometries. Based on the investigation results, we designed the F-Learn strategy to efficiently learn holistic geometry clues in low-level feature maps for structured reconstruction.
    

	\section{Method}
	\label{part:methodology}
	Geometric features extracted at the early convolutional stage fundamentally support the structure reconstruction with precise geometry localization clues. Therefore, we are interested in addressing a key problem for learning reliable low-level geometric features: the low efficiency of fusing geometric fragments in shallow convolution layers. Considering that geometric information is generally high-frequency signals, we are motivated to study the problem mentioned before with frequency analysis for structure reconstruction. In the following of this section, we first present the preliminaries of frequency analysis. Next, we validate the high efficiency of learning in the frequency domain for holistic geometry recovery. Finally, we introduce our frequency-domain feature learning strategy (F-Learn) and the application of the F-Learn in a given structure reconstruction model.
	
	\subsection{Preliminaries of Frequency Analysis}
	The discrete Fourier transform (DFT) is necessary for analyzing image data in the frequency domain. With $I(x,y)$ denoting the color signal at the spatial position $\left(x,y\right)$ of an $M\times N$ image, the DFT converts the information of $I$ to a frequency map $F$ in the frequency domain by~\cref{eq:fourier}.
	\begin{equation}
		\label{eq:fourier}
		F(u, v)=\sum_{x=0}^{M-1} \sum_{y=0}^{N-1} I(x, y) e^{-j 2 \pi(u x / M+v y / N)}.
	\end{equation}
	
	In~\cref{eq:fourier}, $F(u, v)$ is a value that encodes a two-dimension signal changing that belongs to the frequency determined by $u$ and $v$. The frequency map excels in holistically representing signals that show a similar frequency pattern. Therefore, we assume that the DFT is a powerful tool for efficiently learning structure-related geometry features. To validate this hypothesis, we conducted experiments with a geometry recovery task, and analyze the results quantitatively and qualitatively as below.
	
	\subsection{Learning Efficiency Analysis}
	\label{sec:tinyexp}
\begin{figure}[h]
		\centering
		\includegraphics[width=\linewidth]{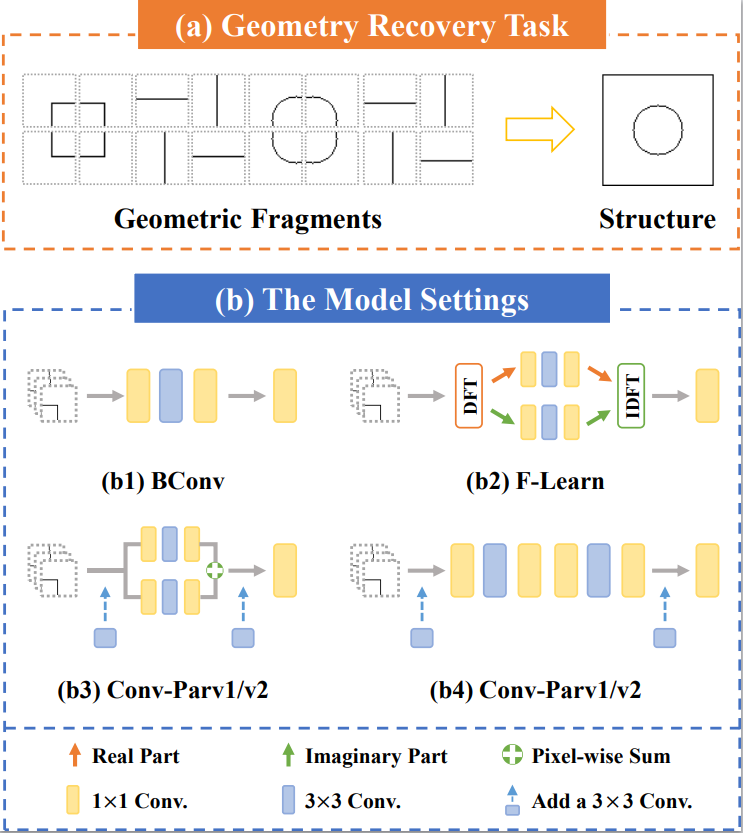}
		\caption{The settings of the simple geometry recovery task. (a) The task of recovering the holistic structure with geometric fragments. (b) The detailed settings of compared models.}
		\label{fig:tinyexp}
	\end{figure}

	In this section, we study the learning efficiency of fusing geometric fragments in the space and frequency domains with a simple geometry recovery task. 
     We first decomposed a geometric structure composed of a circle and a square into several geometric fragments as illustrated in~\cref{fig:tinyexp} (a), and left overlapping between neighboring fragments to avoid trial solutions. 
     
     With the purpose to recover the binary image of the original structure with the geometric fragments, we studied the learning efficiency of frequency-domain and space-domain convolution models. We constructed a frequency-domain learning model (F-Learn) and several space-domain convolution models. The baseline was set as a space-domain convolution module (BConv) built with two $1\times 1$ convolution layers and one $3\times 3$ layer in between. The F-Learn model consists of three key components: (1) a DFT; (2) two parallel convolution modules same with the baseline for separately processing the real and imaginary parts; and (3) an inverse DFT (IDFT) followed by magnitude computation of complex values. For a fair comparison, we also combined a pair of baseline modules into two kinds of space-domain learning models with parallel and cascade structures, namely Conv-Casv1 and Conv-Parv1. Extra $3\times 3$ convolutions were also added into the cascade and parallel space-domain models to simulate the function of DFT and IDFT. The modified models were named Conv-Casv2 and Conv-Parv2 for simplicity. For all models, the channel number of the intermediate feature maps was 64, and a $1\times 1$ convolution was used as the final binary classifier. Detailed model settings are depicted in~\cref{fig:tinyexp}(b). 
	
\begin{figure}[h]
		\centering
        \includegraphics[width=\linewidth]{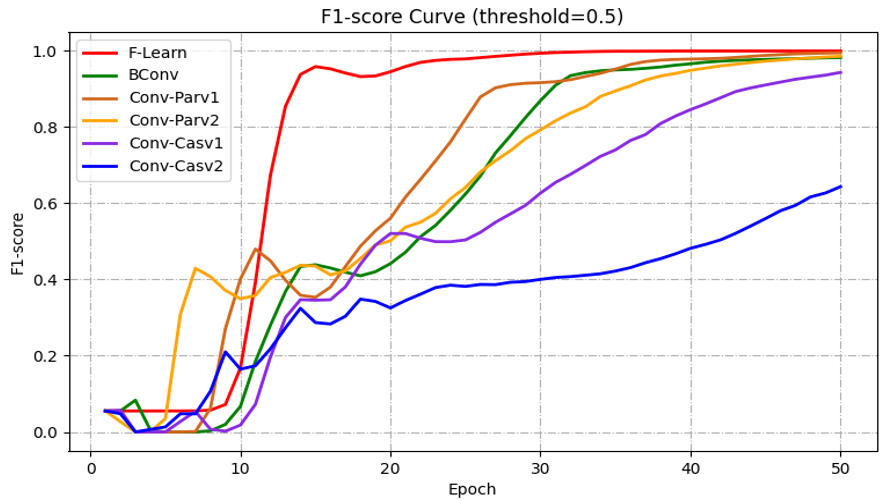}
		\caption{The averaged F1-score curves of $100$ trials for the F-Learn model and the compared space-domain methods.}
		\label{fig:num-tinyexp}
	\end{figure}

\begin{figure}[h]
		\centering
  \includegraphics[width=\linewidth]{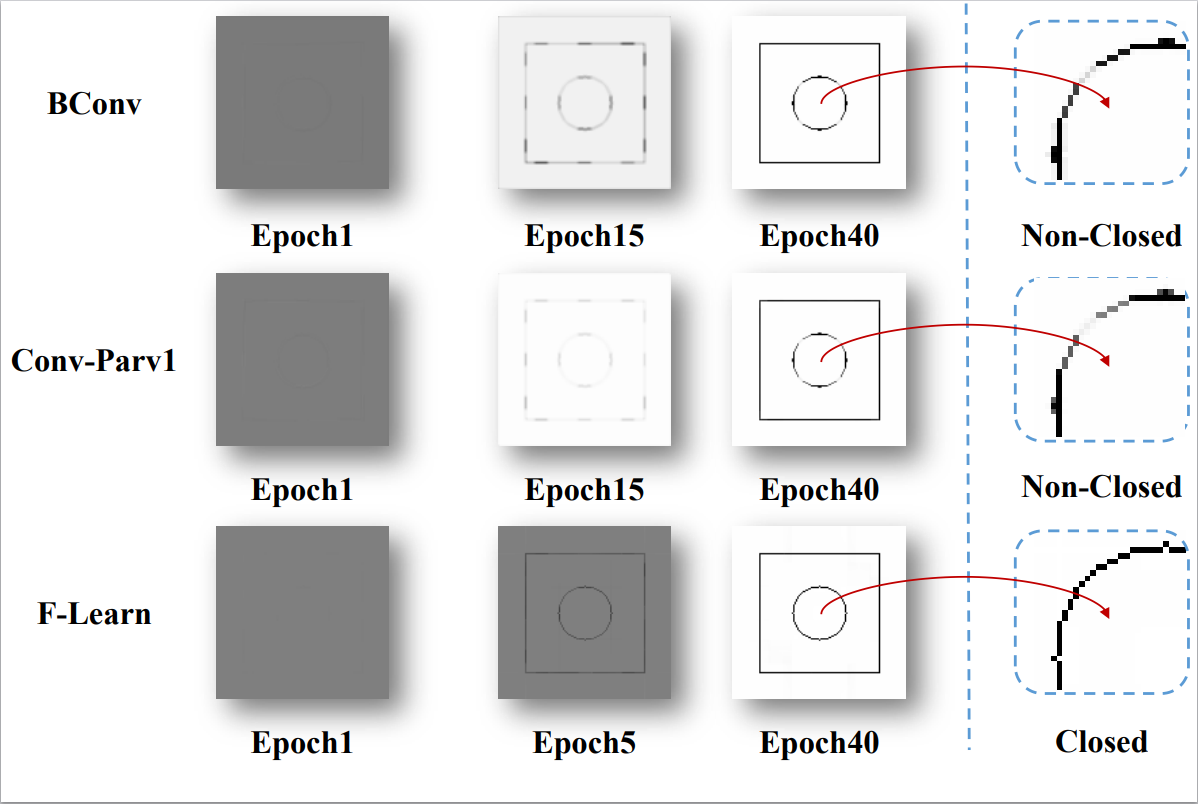}
		\caption{The visualized results of F-Learn, Bconv and Cov-Parv1 at different epochs.}
		\label{fig:vis-tinyexp}
 \end{figure}

	The experiments were conducted on $128\times 128$ binary images of the geometric structure and fragments. The training of all models was completed with an Adam optimizer. The learning rate and training period were set to $0.1$ and 50 epochs, respectively. The binary cross entropy loss (BCELoss) was chosen as the training guide. To alleviate the influence of randomness, we generated $100$ seeds to set $100$ trials, and all models were trained with the same seed in every trial. The performance of all models was evaluated with an F1-score measurement at a threshold of 0.5. Quantitative and qualitative results are shown in~\cref{fig:num-tinyexp} and ~\cref{fig:vis-tinyexp}.
	
	\cref{fig:num-tinyexp} draws the averaged F1-score curves of $100$ trials. Compared with the space-domain counterparts, the frequency-domain model (F-Learn) shows superiority in recovering the complete structure in a more efficient manner.
	
	The visualized results in \cref{fig:vis-tinyexp} also show that the frequency-domain learning model exceeds the basic spatial convolution model with faster convergence speed and better structure recovery quality. 
	
	\cref{fig:vis-tinyexp} reveals that the F-Learn model easily rebuilds the original structure, while the space-domain methods encounter difficulties in efficiently reconstructing the circle. We also find the difference in learning tendency between the F-Learn and space-domain models. The F-Learn model prioritizes holistic integration before further enlarging the gap between the foreground and background. In contrast, the space-domain models first generate high responses to local areas that have strong signal intensity in the input. Hence, the space-domain models prefer overlapped areas without considering structural continuity. In practice, the holistic learning pattern of F-Learn will benefit the topology inference in two ways: (i) holistic geometry clues are ready for learning high-level features at the early stage, which\textbf{ alleviates the burden on learning structure-related low-level features with far supervision signal}; (ii) the holistic geometry learning excels in \textbf{preserving the connection information} in the high-resolution low-level feature maps, which are often reused to \textbf{provide accurate structure localization} for topology inference.  
	
	We attribute the advantages of the frequency domain based model to its ability to holistically process signals that belong to similar frequencies. Based on the findings above, we designed a frequency-domain feature learning strategy to generate holistic geometric clues in the low-level feature maps for structured reconstruction.
	
	\subsection{Frequency-domain Learning Strategy}
	The proposed frequency-domain learning strategy (F-learn) inherits the architecture shown in~\cref{fig:tinyexp}. As the F-Learn strategy will face more complex data and larger batch size in practice than in~\cref{sec:tinyexp}, we replace each convolution layer between DFT and IDFT with a combination (CBR) of a convolution layer, a batch normalization operation, and a relu activation function. We refer to all operations between DFT and IDFT collectively as a CBR group (CBRG). Given a set of geometric feature maps$\{f_i|i=0,...,N\}$, the F-Learn strategy is formulated as
	\begin{equation}
		\centering
		\label{eq:f-learn}
		\begin{aligned}	
			(F^\text{re}_i, F^\text{im}_i) &= \text{DFT}(f_i);\\
			\{\hat{F}^\text{re}_i\}, \{\hat{F}^\text{im}_i\} = \text{CBRG}&(\{F^\text{re}_i\}), \text{CBRG}(\{F^\text{im}_i\});\\
			\{\hat{f}_i\} = |\text{ID}&\text{FT}(\{\hat{F}^\text{re}_i\}, \{\hat{F}^\text{im}_i\})|.
		\end{aligned}
	\end{equation}
	
	The F-learn strategy first converts every $f_i$ into a pair of maps that record the real and imaginary parts of the DFT, i.e., $F^\text{re}_i$ and $F^\text{im}_i$. The maps of the real part $\{F^\text{re}_i\}$ are subsequently processed by a CBRG that contains two $1\times 1$ layers with a $3\times 3$ one in between, and the same operation is also applied to $\{F^\text{im}_i\}$. In the frequency domain, the $1\times 1$ convolution directly fuses the information of the same frequency across the channel dimension, while the $3\times 3$ counterpart enhances features with signals that have similar frequencies. The high-frequency geometric fragments in $\{f_i\}$ are easily combined and enhanced in such a holistic way. The structurally informative space-domain features $\{\hat{f}_i\}$ can be finally obtained through an IDFT along with magnitude computation $|\cdot|$. With the parsimonious design, the F-learn strategy is readily inserted into a hierarchical convolutional backbone for holistic geometry feature learning. 
	
	For feature propagation, the $\{f_i\}$ and $\{\hat{f}_i\}$ are fused into $\{\tilde{f}_i\}$ via
	\begin{equation}
		\{\tilde{f}_i\} = \text{C}_{1\times 1}\text{BR}(\text{Concat}(\{f_i\}, \{\hat{f}_i\})),
	\end{equation}
	where $\text{Concat }$and $\text{C}_{1\times 1}\text{BR}$ denote the concatenation operation and a $1\times 1$ convolution based CBR, respectively.

	The enhanced geometric features $\{\tilde{f}_i\}$ are then fed into the next convolution stage for higher-level feature learning and used to offer precise geometric localization for later structure inference.
	
	The F-Learn strategy can be easily inserted into a convolution base backbone. With the commonly-used ResNet50 backone~\cite{he2016deep} as an example, the F-Learn strategy can be directly deployed after the first convolution to holistically learn geometric features. In experiments, we implemented our F-Learn strategy in a state-of-the-art approach, i.e., the holistic edge attention transformer (HEAT)~\cite{chen2022heat}, for structured reconstruction.

\section{Experiments}
\label{sec:experiments}

\subsection{Experiment Settings}
\subsubsection{Dataset}
We tested our method on a dataset introduced by~\cite{nauata2020vectorizing} for vectorizing world buildings. This dataset is built on the SpaceNet dataset~\cite{van2018spacenet} with 2001 images annotated with roof planar graphs. Each image contains a building instance, and the image is processed into a size of $256 \times 256$. In line with the prior arts~\cite{chen2022heat}, the training/validation/testing split is set to 1601/50/350.

\subsubsection{Implementation Details}
We evaluated the F-learn strategy with the HEAT model for structured reconstruction. The HEAT model was built on a ResNet50 backbone pretrained on ImageNet~\cite{deng2009imagenet}. We performed our experiments with Pytorch~\cite{steiner2019pytorch:} in Python3.7 and used a workstation with one NVIDIA RTX 3090 GPU. We adopted the same settings of loss functions as the HEAT model. The training of the model was completed with an AdamW optimizer, and the training period lasts $800$ epochs. The learning rate was initialized as $2e-4$ and multiplied by a factor of $0.1$ for the last $25\%$ epoch. In line with previous works~\cite{chen2022heat, nauata2020vectorizing}, we used precision, recall, and F1 score to evaluate the quality of structure reconstruction in terms of corner detection, edge inference, and region recovery. 

\begin{table*}[t]
	\centering
	\caption{\textbf{Quantitative comparison between the F-Learn strategy and other methods in terms of corners detection, edge extraction, and region reconstruction.} Prec and F1 are the abbreviations of the precision and f1-score metrics. The higher the scores are, the better the performance is. The best results are marked \first{bold}. (unit:\%)
	}
	{
		\tabcolsep 5pt
		\begin{tabular}{l @{\quad\;\;\;\;}  ccccccccccc@{\;}}
			\addlinespace
			\toprule

			\multicolumn{1}{l}{Evaluation Type $\rightarrow$} &  \multicolumn{1}{c}{} & \multicolumn{1}{c}{} & \multicolumn{3}{c}{Corner} &  \multicolumn{3}{c}{Edge} &  \multicolumn{3}{c}{Region} \\
			\cmidrule(lr){4-6} \cmidrule(lr){7-9} \cmidrule(lr){10-12}
			{Method} & Fully-neural & Joint & Prec & Recall & F1 & Prec & Recall & F1 & Prec & Recall & F1  \\
   
			\midrule
			IP~\cite{nauata2020vectorizing} & - & - & - & - & 74.5 & - & - & 53.1 & - & - & 55.7 \\
			Exp-Cls~\cite{zhang2021structured} & - & - &  92.2 & 75.9 & 83.2 & 75.4 & 60.4 & 67.1 & 74.9 & 54.7 & 63.5 \\
			ConvMPN~\cite{zhang2020conv} & $\checkmark$ & - & 78.0 & 79.7 & 78.8 & 57.0 & 59.7 & 58.1 & 52.4 & 56.5 & 54.4  \\ 
			HAWP~\cite{xue2020holistically} & $\checkmark$ & $\checkmark$ & 90.9 & 81.2 & 85.7 & 76.6 & 68.1 & 72.1 & 74.1 & 55.4 & 63.4 \\
			LETR~\cite{xu2021line} & $\checkmark$ & $\checkmark$ & 87.8 & 74.8 & 80.8 & 59.7 & 58.6 & 59.1 & 68.3 & 48.7 & 56.8 \\
			HEAT~\cite{chen2022heat} & $\checkmark$ & $\checkmark$ & 91.7 & 83.0 & 87.1 & 80.6 & 72.3 & 76.2 & 76.4 & 65.6 & 70.6 \\
			HEAT(retrain)~\cite{chen2022heat} & $\checkmark$ & $\checkmark$ & 91.6 & 83.0 & 87.1 & 80.3 & 72.4 & 76.1 & 75.5 & 65.3 & 70.0 \\
			\midrule
			\textbf{F-Learn (Ours)} & $\checkmark$ & $\checkmark$ & \first{93.2} & \first{84.4} & \first{88.6} & \first{83.6} & \first{75.0} & \first{79.1} & \first{79.5} & \first{68.1} & \first{73.4} \\ 
			\bottomrule
		\end{tabular}
	}
	\label{tab:num-com}
\end{table*}

\subsubsection{Competing Methods}
We compare the F-Learn strategy with six methods: HEAT~\cite{chen2022heat}, ConvMPN~\cite{zhang2020conv}, IP~\cite{nauata2020vectorizing}, Exp-cls~\cite{zhang2021structured}, HAWP~\cite{xue2020holistically} and LETR~\cite{xu2021line}. 

\textbf{HEAT} is an attention-based method that takes a 2D raster image as an input and reconstructs a planar graph in an end-to-end manner. HEAT works via three steps: 1) extracting hierarchical features by ResNet50; 2) detecting corners with multi-scale features enriched by a deformable attention module; and 3) employing two weight-sharing Transformer decoders to classify edges and reason structures with detected corners. Our F-Learn strategy is adopted in the first step for learning structure-related features.

Conv-MPN uses a graph neural network to infer edges with a pre-trained corner detector. IP firstly detects geometric primitives and then reconstructs a planar graph through several post-processing steps. Exp-cls is based on the geometric primitives produced by other methods (e.g., IP and Conv-MPN) and reconstructs a planar graph through an explore-and-classify framework. LCNN and HAWP are methods specifically proposed for wireframe parsing. LETR is a transformer-based method that directly generates lines without post-processing and heuristic guidance.

\subsection{Comparison and Analysis}
\subsubsection{Quantitative Analysis}
\soulregister{\cref}7
The quantitative results displayed in~\cref{tab:num-com} show that our F-Learn strategy advances roof structure reconstruction with state-of-the-art performance. The F-Learn strategy edges out other compared methods by significant performance gains regarding corner detection, edge inference, and region reconstruction under all metrics. 

With respect to the detection of corners and edges, the F-Learn strategy outperforms the second-best by at least 1.0\% and 3.0\% in the precision measurement, while 1.4\% and 2.6\% under the recall metric. The improvement in extracting low-level geometric primitives indicates the outstanding ability of the F-Learn strategy for providing holistic and accurate geometric clues. The F-Learn strategy also brings at least a 2.8\% gain in F1 score in terms of region reconstruction, which reveals the assistance of the F-Learn strategy in inferring correct topological relations. Besides, it is worth noticing that the F-Learn strategy significantly outperforms the original HEAT model under all metrics, but only brings a little computation increase compared to HEAT. ~\cref{tab:compute-increase} presents a detailed comparison of computation requirements.


\begin{table}[h]
	\centering
 
    \caption{\textbf{Quantitative evaluation of the change of computing efficiency brought by F-Learn.} Param. and FLOPs are the abbreviations of the parameter quantity and floating point operations, used to measure the complexity of an algorithm/model. Time per Image represents the time it takes the model to infer an image and is used to evaluate the calculation speed of the model.
	}
	        \vspace{0.3em}
	{
		\tabcolsep 5pt
		\begin{tabular}{lccc}
			\addlinespace
			\toprule
			& Param. & FLOPs & Time per Image \\
			\toprule
			  w/o F-Learn & 68.747M & 670.919G & 0.10s \\
			  w/ F-Learn & 68.847M & 684.039G & 0.11s \\
			\bottomrule
		\end{tabular}
	}
	\label{tab:compute-increase}
\end{table}

In~\cref{tab:compute-increase}, the computation increase of F-Learn strategy in terms of parameters, FLOPs, and inference time per image can be clearly seen, where the amount is little. Especially, the parameter increase brought by F-Learn strategy is only 0.145$\%$, which is almost negligible. The results further demonstrates that the F-Learn strategy is simple yet effective.

\begin{figure*}[h]
	\centering
	\includegraphics[width=\linewidth]{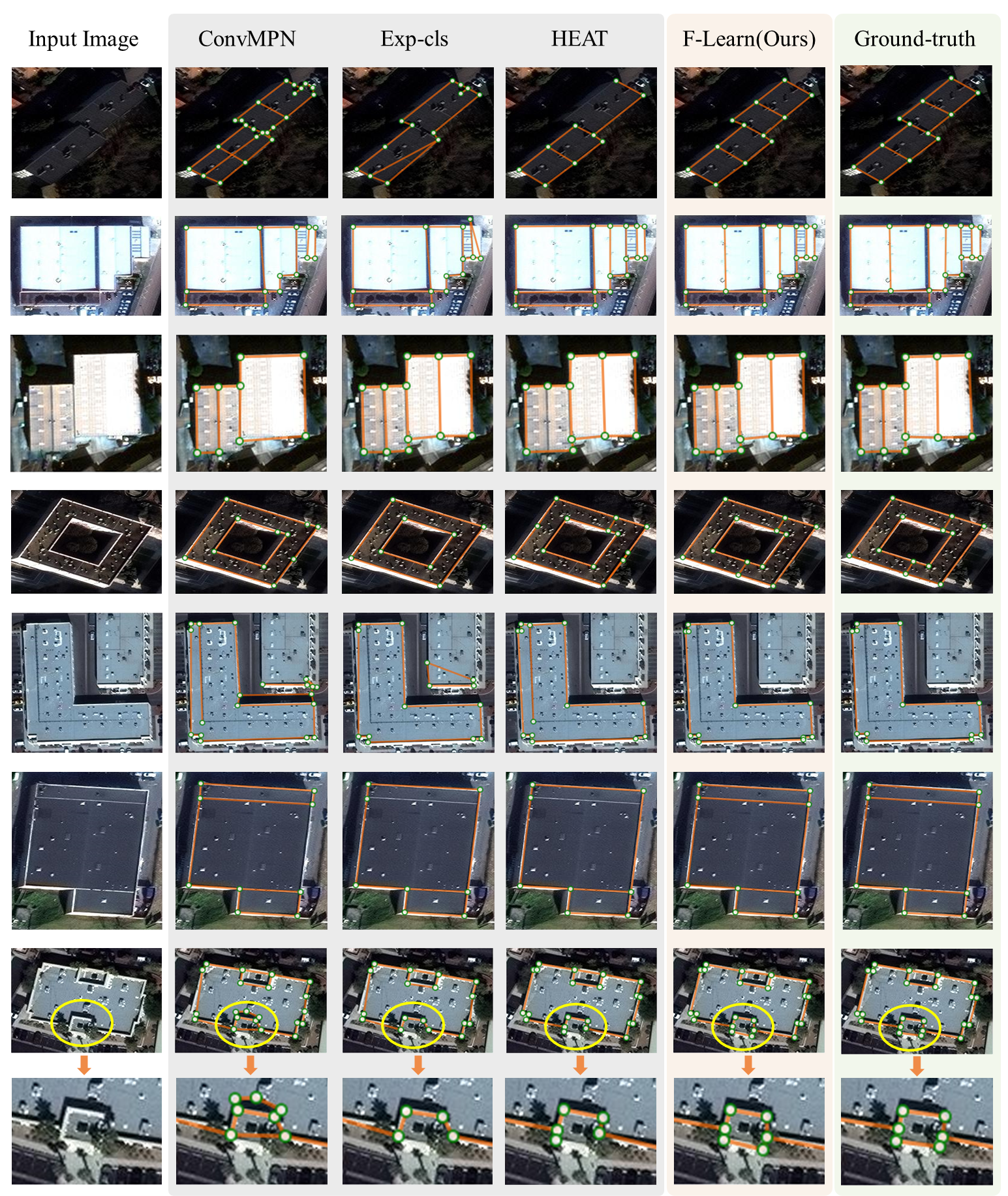}
	\vspace{-1em}
	\caption{\textbf{Qualitative results on outdoor roof structure reconstruction.} Other methods build wrong topological relations due to redundant or missing geometric primitives. Our F-Learn strategy derives the correct topological principles with corners and edges more fitted to the holistic structures.}
	\label{fig:vis-comp}
\end{figure*}

\subsubsection{Qualitative Analysis}
For perceptual comparison, we also present the visualized reconstruction results in~\cref{fig:vis-comp}. Compared to other methods, our F-Learn is capable of detecting low-level geometric primitives that are consistent with the holistic structure. 

When reconstructing the roof shown in the first row, ConvMPN and Exp-cls tend to detect redundant corners due to the shadow inference and therefore wrongly separate the roof regions. The original HEAT outputs a non-closed roof structure because key edges are missing. In contrast, the corners and edges learned with the F-Learn strategy are better fitted into the overall structure. This is because the F-Learn strategy can offer low-level features that are rich in holistic geometric clues for topology-preserved roof structure reconstruction. The results in the following rows additionally validate that the F-Learn strategy can effectively extract geometric primitives that are easily ignored by other methods, especially the original HEAT model. By comparing the inferred planar graphs in the fifth row, one can also find that the F-Learn improves the HEAT model with robustness to irrelevant structure information.

\subsection{Ablation Studies}
\subsubsection{Comparison with Space-domain Methods}
\label{sec:diff-conv}
In addition to the exploration with a simple geometry fusion task in~\cref{sec:tinyexp}, we further study the F-Learn strategy with the space-domain learning strategies in the real scene for roof structure reconstruction. Similar to~\cref{sec:tinyexp}, we construct a space-domain learning baseline (BConv) with a basic module composed of two $1\times 1$ convolution layers and one $3\times 3$ layer in between. Furthermore, we design two more space-domain strategies with the basic modules arranged in parallel and cascade, denoted as Conv-Parv. and Conv-Casv., respectively. Same with the F-Learn, additional operations of batch normalization and ReLU activation are added after each convolution layer used in every space-domain strategy. 

\begin{table}[h]
	\centering
    \caption{\textbf{Quantitative comparisons with space-domain learning methods.} The F-Learn strategy is compared with the space-domain counterparts under the F1-score measurement in terms of corner detection, edge inference, and region reconstruction. The higher the scores are, the better the performance is. The best results are marked \first{bold}. (unit:\%)
	}
	        \vspace{0.3em}
	{
		\tabcolsep 5pt
		\begin{tabular}{l@{\quad\;\;\;\;} ccc}
			\addlinespace
			\toprule
			Method & Corner &  Edge &  Region \\
			\midrule
			HEAT & 87.1 & 76.1 & 70.0 \\
			HEAT-BConv & 87.4 & 76.2 & 70.7 \\
			HEAT-Conv-Cas. & 88.1 & 77.1 & 70.2 \\
			HEAT-Conv-Par. & 87.7 & 76.8 & 69.5 \\
			\midrule
			\textbf{F-Learn} & \first{88.6} & \first{79.1} & \first{73.4} \\
			\bottomrule
		\end{tabular}
	}
	\label{tab:diff-conv}
\end{table}

With the F1-score measurement,~\cref{tab:diff-conv} presents the performance of the F-learn and space-domain learning strategies in detecting geometric primitives. Although some space-domain methods are comparable to the proposed F-Learn strategy in corner extraction, it can be seen that the F-Learn strategy gains over these methods in detecting edges and regions by at least 2\% and 2.7\%, respectively. Because edges and regions are primitives closely related to roof topology, this phenomenon demonstrates the holistic geometry learned by the F-Learn is beneficial for the topology inference for roof reconstruction.

\subsubsection{F-Learn Strategy at Different Layers}
We investigate the effect of applying the F-Learn strategy to feature maps of different levels with the HEAT model and the hierarchical ResNet50 backbone. The F-learn strategy is deployed to process feature maps generated at the first convolution layer, the first residual learning stage, and the second one. The sizes of the corresponding feature maps are $128\times 128$, $64×\times 64$, and $32\times 32$, respectively. The quantitative results are presented in~\cref{tab:diff-layer}.
\begin{table}[h]
	\centering
    \caption{\textbf{Quantitative comparison of applying the F-Learn strategy to features at different levels.} L0: the first convolution layer; L1: the first residual learning stage; L2: the second residual learning stage. Size: the feature map size. The F1-score metric is used for evaluating the performance of corner detection, edge inference, and region recovery. Higher scores mean better results. The \first{bold} values represent the best performance. (unit:\%)
	}
	\vspace{0.3em}
	{
		\tabcolsep 5pt
		\begin{tabular}{l@{\quad\;\;\;\;} cccc}
			\addlinespace
			\toprule
			Method & Size & Corner &  Edge &  Region \\
			\midrule
			HEAT & - & 87.1 & 76.1 & 70.0 \\
			F-Learn-L0 & $128\times 128$ & \first{88.6} & \first{79.1} & \first{73.4} \\
			F-Learn-L1 & $64\times 64$ & 87.7 & 76.4 & 70.9 \\
			F-Learn-L2 & $32\times 32$  & 86.7 & 75.3 & 69.0 \\
			\bottomrule
		\end{tabular}
	}
	\label{tab:diff-layer}
\end{table}

\cref{tab:diff-layer} implies that the higher the resolution is, the better the F-Learn strategy works. It is because high-resolution low-level feature maps contain more high-frequency geometric clues, while the low-resolution high-level ones are richer in abstract semantic information but poorer in geometric details. Therefore, the high-resolution feature maps are more suitable than the low-resolution counterparts for the F-Learn strategy to learn structurally informative geometries in roof reconstruction.

\begin{table*}[h]
	\centering
	\caption{\textbf{Quantitative comparison of F-Learn strategy with HEAT built on different backbones, including ResNet18, ResNet34, and ResNet50}. Prec and F1 are the abbreviations of the precision and f1-score metrics. The higher the scores are, the better the performance is. The best results are marked \first{bold}. (unit:\%)
	}
         \vspace{0.3em}
	{
		\tabcolsep 5pt
		\begin{tabular}{lcccccccccc}
			\addlinespace
			\toprule
			 &  & \multicolumn{3}{c}{Corner} &  \multicolumn{3}{c}{Edge} &  \multicolumn{3}{c}{Region} \\
			\cmidrule(lr){3-5} \cmidrule(lr){6-8} \cmidrule(lr){9-11}
			& Backbone &  Prec & Recall & F1 & Prec & Recall & F1 & Prec & Recall & F1  \\
			\midrule
			  w/o F-Learn & ResNet18 & 91.6 & 81.9 & 86.5 & 79.9 & 70.1 & 74.7 & 76.3 & 61.6 & 68.2\\
			  w/ F-Learn & ResNet18 & \textbf{91.7} & \textbf{82.2} & \textbf{86.7} & \textbf{80.8} & \textbf{70.4} & \textbf{75.2} & \textbf{77.1} & \textbf{62.9} & \textbf{69.3} \\
                \midrule
			  w/o F-Learn & ResNet34 & 90.5 & 82.6 & 86.4 & 78.3 & 71.1 & 74.5 & 73.4 & 63.4 & 68.0 \\
			  w/ F-Learn & ResNet34 & \textbf{92.1} & \textbf{83.2} & \textbf{87.4} & \textbf{80.2} & \textbf{71.8} & \textbf{75.8} & \textbf{76.2} & \textbf{64.1} & \textbf{69.6} \\
                \midrule
			  w/o F-Learn & ResNet50 & 91.6 & 83.0 & 87.1 & 80.3 & 72.4 & 76.1 & 75.5 & 65.3 & 70.0 \\
            w/ F-Learn & ResNet50 & \textbf{93.2} & \textbf{84.4} & \textbf{88.6} & \textbf{83.6}& \textbf{75.0} & \textbf{79.1} & \textbf{79.5} & \textbf{68.1} & \textbf{73.4} \\
			\bottomrule
		\end{tabular}
	}
	\label{tab:diff-backbone}
\end{table*}

\subsubsection{F-Learn Strategy with Various Backbones}

To further demonstrate the effectiveness of the F-learn strategy, we evaluate the performance improvement brought by the F-Learn strategy with different backbones. The quantitative results are presented in~\cref{tab:diff-backbone}. As shown by~\cref{tab:diff-backbone}, with the deployment of the F-Learn strategy, the performance of HEAT is consistently improved in terms of corners detection, edge extraction, and region reconstruction. 

\subsubsection{F-Learn Strategy in Topology Inference}
In this section, we further explore the effectiveness of the F-Learn strategy in topology inference. Edges and regions are primitives that highly relate to the inference of topological relationships, and these primitives are learned with corners and low-level features in the HEAT model. Therefore, we directly use the ground-truth corner map to focus on topology inference. We compare our F-Learn strategy with the space-domain methods used in~\cref{sec:diff-conv}.

\begin{table}[h]
	\centering
        \caption{\textbf{Quantitative results of the F-Learn and space-domain strategies with respect to topology inference.} The HEAT method with corner ground truth is set as the baseline. With corner annotations, all methods are studied with a focus on topology inference. The higher scores mean better results, and the best performance is \first{bolded}. (unit: \%)
	}
        \vspace{0.3em}
	\resizebox{\linewidth}{!}
	{
		\tabcolsep 5pt
		\begin{tabular}{l@{\quad\;\;\;\;} cccccc}
			\addlinespace
			\toprule
			\multicolumn{1}{l}{Evaluation Type $\rightarrow$} &  \multicolumn{3}{c}{Edge} &  \multicolumn{3}{c}{Region} \\
			\cmidrule(lr){2-4} \cmidrule(lr){5-7}
			{Method} & Prec & Recall & F1 & Prec & Recall & F1  \\
			\midrule
			HEAT & 93.5 & 84.1 & 88.5 & 90.2 & 73.8 & 81.2 \\
			HEAT-BConv & 94.4 & 83.9 & 88.8 & 89.4 & 71.7 & 79.6 \\
			HEAT-Conv-Cas & 94.8 & 85.3 & 89.8 & \first{91.3} & 73.7 & 81.6 \\
			HEAT-Conv-Par  & 93.8 & 84.8 & 89.1 & 89.7 & 73.9 & 81.0 \\
			\midrule
			\textbf{F-Learn} & \first{95.1} & \first{87.5} & \first{91.1} & 89.5 & \first{77.6} & \first{83.1} \\ 
			\bottomrule
		\end{tabular}
	}
	\label{tab:gt-corner}
\end{table}

\cref{tab:gt-corner} presents the numeric results, and it can be seen that our F-Learn strategy brings the highest gains in most metrics for edge inference and region reconstruction. This phenomenon demonstrates that the F-Learn strategy effectively supports topology inference with an outstanding ability for holistic geometry learning.

\section{Conclusions}
In this paper, we present a frequency-domain feature learning strategy (F-Learn) to tackle the issue of wrong topology recovery caused by the lack of holistic clues in low-level features. Experiments with a geometry recovery task convincingly validate the efficiency of the F-Learn strategy in learning holistic geometry. In terms of the real scene, the F-Learn strategy achieves significant performance improvement of topological principle inference for roof structure reconstruction. The ablation studies verify that the F-Learn strategy outperforms the space-domain learning counterparts in capturing holistic geometric features regarding complex real scenes. We believe that it is promising to further explore the holistic learning ability brought by frequency analysis in more vision tasks.

\section{Acknowledgement}
This research is supported by NSFC-projects under Grant 42071370, the Fundamental Research Funds for the Central Universities of China under Grant 2042022dx0001, and the Open fund of Key Laboratory of Urban Land Resources Monitoring and Simulation, Ministry of Natural Resources under Grant KF202106084.


{\small
	\bibliographystyle{ieee_fullname}
	\bibliography{egbib}

\begin{thebibliography}{10}
\providecommand{\url}[1]{#1}
\csname url@samestyle\endcsname
\providecommand{\newblock}{\relax}
\providecommand{\bibinfo}[2]{#2}
\providecommand{\BIBentrySTDinterwordspacing}{\spaceskip=0pt\relax}
\providecommand{\BIBentryALTinterwordstretchfactor}{4}
\providecommand{\BIBentryALTinterwordspacing}{\spaceskip=\fontdimen2\font plus
\BIBentryALTinterwordstretchfactor\fontdimen3\font minus
  \fontdimen4\font\relax}
\providecommand{\BIBforeignlanguage}[2]{{%
\expandafter\ifx\csname l@#1\endcsname\relax
\typeout{** WARNING: IEEEtran.bst: No hyphenation pattern has been}%
\typeout{** loaded for the language `#1'. Using the pattern for}%
\typeout{** the default language instead.}%
\else
\language=\csname l@#1\endcsname
\fi
#2}}
\providecommand{\BIBdecl}{\relax}
\BIBdecl

\bibitem{liu2019richer}
Y.~Liu, M.-M. Cheng, X.~Hu, J.-W. Bian, L.~Zhang, X.~Bai, and J.~Tang, ``Richer
  convolutional features for edge detection,'' \emph{IEEE Trans. Pattern Anal.
  Mach. Intell.}, vol.~41, no.~8, pp. 1939--1946, 2019.

\bibitem{Woo_2018_ECCV}
S.~Woo, J.~Park, J.-Y. Lee, and I.~S. Kweon, ``Cbam: Convolutional block
  attention module,'' in \emph{ECCV}, September 2018.

\bibitem{chen2016attention}
L.-C. Chen, Y.~Yang, J.~Wang, W.~Xu, and A.~L. Yuille, ``Attention to scale:
  Scale-aware semantic image segmentation,'' in \emph{CVPR}, 2016, pp.
  3640--3649.

\bibitem{xu2015show}
K.~Xu, J.~Ba, R.~Kiros, K.~Cho, A.~Courville, R.~Salakhudinov, R.~Zemel, and
  Y.~Bengio, ``Show, attend and tell: Neural image caption generation with
  visual attention,'' in \emph{ICML}.\hskip 1em plus 0.5em minus 0.4em\relax
  PMLR, 2015, pp. 2048--2057.

\bibitem{DFN}
X.~Jia, B.~Brabandere, T.~Tuytelaars, and L.~Van~Gool, ``Dynamic filter
  networks,'' \emph{NeurIPS}, 01 2016.

\bibitem{DFF}
Y.~Hu, Y.~Chen, X.~Li, and J.~Feng, ``Dynamic feature fusion for semantic edge
  detection,'' in \emph{IJCAI}, 08 2019, pp. 782--788.

\bibitem{fram1975on}
J.~R. Fram and E.~S. Deutsch, ``On the quantitative evaluation of edge
  detection schemes and their comparison with human performance,'' \emph{IEEE
  Trans. Comput.}, vol.~24, no.~6, pp. 616--628, 1975.

\bibitem{roberts1963machine}
L.~G. Roberts, ``Machine perception of three-dimensional solids,'' \emph{PhD
  thesis, Massachusetts Institute of Technology}, 1963.

\bibitem{kittler1983accuracy}
J.~Kittler, ``On the accuracy of the sobel edge detector,'' \emph{Image and
  Vision Computing}, vol.~1, no.~1, pp. 37--42, 1983.

\bibitem{milletari2016v}
F.~Milletari, N.~Navab, and S.-A. Ahmadi, ``V-net: Fully convolutional neural
  networks for volumetric medical image segmentation,'' in \emph{3DV}.\hskip
  1em plus 0.5em minus 0.4em\relax IEEE, 2016, pp. 565--571.

\bibitem{davis1975survey}
L.~S. Davis, ``A survey of edge detection techniques,'' \emph{Computer graphics
  and image processing}, vol.~4, no.~3, pp. 248--270, 1975.

\bibitem{gonzales1987digital}
R.~C. Gonzales and P.~Wintz, \emph{Digital image processing}.\hskip 1em plus
  0.5em minus 0.4em\relax Addison-Wesley Longman Publishing Co., Inc., 1987.

\bibitem{Hueckel1971}
\BIBentryALTinterwordspacing
M.~H. Hueckel, ``An operator which locates edges in digitized pictures,''
  \emph{J. ACM}, vol.~18, no.~1, p. 113–125, 1971. [Online]. Available:
  \url{https://doi.org/10.1145/321623.321635}
\BIBentrySTDinterwordspacing

\bibitem{Ferrari2008Groups}
V.~Ferrari, L.~Fevrier, F.~Jurie, and C.~Schmid, ``Groups of adjacent contour
  segments for object detection,'' \emph{IEEE Trans. Pattern Anal. Mach.
  Intell.}, vol.~30, no.~1, pp. p.36--51, 2008.

\bibitem{song2020edgestereo:}
X.~Song, X.~Zhao, L.~Fang, H.~Hu, and Y.~Yu, ``Edgestereo: An effective
  multi-task learning network for stereo matching and edge detection,''
  \emph{International Journal of Computer Vision}, 2020.

\bibitem{kokkinos2015pushing}
I.~Kokkinos, ``Pushing the boundaries of boundary detection using deep
  learning,'' \emph{arXiv preprint arXiv:1511.07386}, 2015.

\bibitem{xie2017hed}
S.~Xie and Z.~Tu, ``Holistically-nested edge detection,'' \emph{International
  Journal of Computer Vision}, vol. 125, no.~1, pp. 3--18, 2017.

\bibitem{Liu2017Object}
J.~Liu, T.~Ren, Y.~Wang, S.~H. Zhong, J.~Bei, and S.~Chen, ``Object proposal on
  rgb-d images via elastic edge boxes,'' \emph{Neurocomputing}, vol. 236, no.
  MAY2, pp. 134--146, 2017.

\bibitem{M2016A}
D.~A. Mély, J.~Kim, M.~Mcgill, Y.~Guo, and T.~Serre, ``A systematic comparison
  between visual cues for boundary detection,'' \emph{Vision Research}, vol.
  120, pp. 93--107, 2016.

\bibitem{amfm_pami2011}
P.~Arbelaez, M.~Maire, C.~Fowlkes, and J.~Malik, ``Contour detection and
  hierarchical image segmentation,'' \emph{IEEE Trans. Pattern Anal. Mach.
  Intell.}, vol.~33, no.~5, pp. 898--916, 2011.

\bibitem{canny1986a}
J.~Canny, ``A computational approach to edge detection,'' \emph{IEEE Trans.
  Pattern Anal. Mach. Intell.}, vol.~8, no.~6, pp. 679--698, 1986.

\bibitem{Wang2018Deep}
Y.~{Wang}, X.~{Zhao}, Y.~{Li}, and K.~{Huang}, ``Deep crisp boundaries: From
  boundaries to higher-level tasks,'' \emph{IEEE Trans. Image Process},
  vol.~28, no.~3, pp. 1285--1298, 2019.

\bibitem{Arbel2011Contour}
P.~Arbel{\'{a}}ez, M.~Maire, C.~Fowlkes, and J.~Malik, ``Contour detection and
  hierarchical image segmentation,'' \emph{IEEE Trans. Pattern Anal. Mach.
  Intell.}, vol.~33, no.~5, pp. 898--916, 2011.

\bibitem{steiner2019pytorch:}
B.~Steiner, Z.~Devito, S.~Chintala, S.~Gross, A.~Paszke, F.~Massa, A.~Lerer,
  G.~Chanan, Z.~Lin, E.~Yang \emph{et~al.}, ``Pytorch: An imperative style,
  high-performance deep learning library,'' in \emph{NeurIPS}, 2019, pp.
  8026--8037.

\bibitem{he2019bi-directional}
J.~He, S.~Zhang, M.~Yang, Y.~Shan, and T.~Huang, ``Bi-directional cascade
  network for perceptual edge detection,'' in \emph{CVPR}, 2019, pp.
  3828--3837.

\bibitem{Bertasius2015DeepEdge}
G.~Bertasius, J.~Shi, and L.~Torresani, ``Deepedge: A multi-scale bifurcated
  deep network for top-down contour detection,'' in \emph{CVPR}, 2015.

\bibitem{Wei2015DeepContour}
W.~Shen, X.~Wang, Y.~Wang, X.~Bai, and Z.~Zhang, ``Deepcontour: A deep
  convolutional feature learned by positive-sharing loss for contour
  detection,'' in \emph{CVPR}, 2015, pp. 3982--3991.

\bibitem{deng2009imagenet}
J.~Deng, W.~Dong, R.~Socher, L.-J. Li, K.~Li, and L.~Fei-Fei, ``Imagenet: A
  large-scale hierarchical image database,'' in \emph{CVPR}, 2009, pp.
  248--255.

\bibitem{deng2018learning}
R.~Deng, C.~Shen, S.~Liu, H.~Wang, and X.~Liu, ``Learning to predict crisp
  boundaries,'' in \emph{ECCV}, 2018, pp. 570--586.

\bibitem{Isola2014Crisp}
P.~Isola, D.~Zoran, D.~Krishnan, and E.~H. Adelson, ``Crisp boundary detection
  using pointwise mutual information,'' in \emph{ECCV}, 2014.

\bibitem{Hallman2015Oriented}
S.~{Hallman} and C.~C. {Fowlkes}, ``Oriented edge forests for boundary
  detection,'' in \emph{CVPR}, 2015, pp. 1732--1740.

\bibitem{law2018cornernet}
H.~Law and J.~Deng, ``Cornernet: Detecting objects as paired keypoints,'' in
  \emph{Proceedings of the European conference on computer vision (ECCV)},
  2018, pp. 734--750.

\bibitem{zhang2020conv}
F.~Zhang, N.~Nauata, and Y.~Furukawa, ``Conv-mpn: Convolutional message passing
  neural network for structured outdoor architecture reconstruction,'' in
  \emph{Proceedings of the IEEE/CVF Conference on Computer Vision and Pattern
  Recognition}, 2020, pp. 2798--2807.

\bibitem{huang2018learning}
K.~Huang, Y.~Wang, Z.~Zhou, T.~Ding, S.~Gao, and Y.~Ma, ``Learning to parse
  wireframes in images of man-made environments,'' in \emph{Proceedings of the
  IEEE Conference on Computer Vision and Pattern Recognition}, 2018, pp.
  626--635.

\bibitem{xu2019frequency}
Z.-Q.~J. Xu, Y.~Zhang, T.~Luo, Y.~Xiao, and Z.~Ma, ``Frequency principle:
  Fourier analysis sheds light on deep neural networks,'' \emph{arXiv preprint
  arXiv:1901.06523}, 2019.

\bibitem{basri2020frequency}
R.~Basri, M.~Galun, A.~Geifman, D.~Jacobs, Y.~Kasten, and S.~Kritchman,
  ``Frequency bias in neural networks for input of non-uniform density,'' in
  \emph{International Conference on Machine Learning}.\hskip 1em plus 0.5em
  minus 0.4em\relax PMLR, 2020, pp. 685--694.

\bibitem{xu2020learning}
K.~Xu, M.~Qin, F.~Sun, Y.~Wang, Y.-K. Chen, and F.~Ren, ``Learning in the
  frequency domain,'' in \emph{Proceedings of the IEEE/CVF Conference on
  Computer Vision and Pattern Recognition}, 2020, pp. 1740--1749.

\bibitem{jiang2021focal}
L.~Jiang, B.~Dai, W.~Wu, and C.~C. Loy, ``Focal frequency loss for image
  reconstruction and synthesis,'' in \emph{Proceedings of the IEEE/CVF
  International Conference on Computer Vision}, 2021, pp. 13\,919--13\,929.

\bibitem{dzanic2020fourier}
T.~Dzanic, K.~Shah, and F.~Witherden, ``Fourier spectrum discrepancies in deep
  network generated images,'' \emph{Advances in neural information processing
  systems}, vol.~33, pp. 3022--3032, 2020.

\bibitem{qin2021fcanet}
Z.~Qin, P.~Zhang, F.~Wu, and X.~Li, ``Fcanet: Frequency channel attention
  networks,'' in \emph{Proceedings of the IEEE/CVF international conference on
  computer vision}, 2021, pp. 783--792.

\bibitem{chen2016compressing}
W.~Chen, J.~Wilson, S.~Tyree, K.~Q. Weinberger, and Y.~Chen, ``Compressing
  convolutional neural networks in the frequency domain,'' in \emph{Proceedings
  of the 22nd ACM SIGKDD international conference on knowledge discovery and
  data mining}, 2016, pp. 1475--1484.

\bibitem{liu2018frequency}
Z.~Liu, J.~Xu, X.~Peng, and R.~Xiong, ``Frequency-domain dynamic pruning for
  convolutional neural networks,'' \emph{Advances in neural information
  processing systems}, vol.~31, 2018.

\bibitem{wang2018packing}
Y.~Wang, C.~Xu, C.~Xu, and D.~Tao, ``Packing convolutional neural networks in
  the frequency domain,'' \emph{IEEE transactions on pattern analysis and
  machine intelligence}, vol.~41, no.~10, pp. 2495--2510, 2018.

\bibitem{huang2021rda}
J.~Huang, D.~Guan, A.~Xiao, and S.~Lu, ``Rda: Robust domain adaptation via
  fourier adversarial attacking,'' in \emph{Proceedings of the IEEE/CVF
  international conference on computer vision}, 2021, pp. 8988--8999.

\bibitem{huang2021fsdr}
------, ``Fsdr: Frequency space domain randomization for domain
  generalization,'' in \emph{Proceedings of the IEEE/CVF Conference on Computer
  Vision and Pattern Recognition}, 2021, pp. 6891--6902.

\bibitem{tancik2020fourier}
M.~Tancik, P.~Srinivasan, B.~Mildenhall, S.~Fridovich-Keil, N.~Raghavan,
  U.~Singhal, R.~Ramamoorthi, J.~Barron, and R.~Ng, ``Fourier features let
  networks learn high frequency functions in low dimensional domains,''
  \emph{Advances in Neural Information Processing Systems}, vol.~33, pp.
  7537--7547, 2020.

\bibitem{misra2021end}
I.~Misra, R.~Girdhar, and A.~Joulin, ``An end-to-end transformer model for 3d
  object detection,'' in \emph{Proceedings of the IEEE/CVF International
  Conference on Computer Vision}, 2021, pp. 2906--2917.

\bibitem{barron2021mip}
J.~T. Barron, B.~Mildenhall, M.~Tancik, P.~Hedman, R.~Martin-Brualla, and P.~P.
  Srinivasan, ``Mip-nerf: A multiscale representation for anti-aliasing neural
  radiance fields,'' in \emph{Proceedings of the IEEE/CVF International
  Conference on Computer Vision}, 2021, pp. 5855--5864.

\bibitem{duan2019centernet}
K.~Duan, S.~Bai, L.~Xie, H.~Qi, Q.~Huang, and Q.~Tian, ``Centernet: Keypoint
  triplets for object detection,'' in \emph{Proceedings of the IEEE/CVF
  international conference on computer vision}, 2019, pp. 6569--6578.

\bibitem{huan2021unmixing}
L.~Huan, N.~Xue, X.~Zheng, W.~He, J.~Gong, and G.-S. Xia, ``Unmixing
  convolutional features for crisp edge detection,'' \emph{IEEE Transactions on
  Pattern Analysis and Machine Intelligence}, vol.~44, no.~10, pp. 6602--6609,
  2021.

\bibitem{detone2018superpoint}
D.~DeTone, T.~Malisiewicz, and A.~Rabinovich, ``Superpoint: Self-supervised
  interest point detection and description,'' in \emph{Proceedings of the IEEE
  conference on computer vision and pattern recognition workshops}, 2018, pp.
  224--236.

\bibitem{duda1972use}
R.~O. Duda and P.~E. Hart, ``Use of the hough transformation to detect lines
  and curves in pictures,'' \emph{Communications of the ACM}, vol.~15, no.~1,
  pp. 11--15, 1972.

\bibitem{lin2020deep}
Y.~Lin, S.~L. Pintea, and J.~C. van Gemert, ``Deep hough-transform line
  priors,'' in \emph{Computer Vision--ECCV 2020: 16th European Conference,
  Glasgow, UK, August 23--28, 2020, Proceedings, Part XXII 16}.\hskip 1em plus
  0.5em minus 0.4em\relax Springer, 2020, pp. 323--340.

\bibitem{zhang2019ppgnet}
Z.~Zhang, Z.~Li, N.~Bi, J.~Zheng, J.~Wang, K.~Huang, W.~Luo, Y.~Xu, and S.~Gao,
  ``Ppgnet: Learning point-pair graph for line segment detection,'' in
  \emph{Proceedings of the IEEE/CVF Conference on Computer Vision and Pattern
  Recognition}, 2019, pp. 7105--7114.

\bibitem{xue2019learning}
N.~Xue, S.~Bai, F.~Wang, G.-S. Xia, T.~Wu, and L.~Zhang, ``Learning attraction
  field representation for robust line segment detection,'' in
  \emph{Proceedings of the IEEE/CVF Conference on Computer Vision and Pattern
  Recognition}, 2019, pp. 1595--1603.

\bibitem{xu2021line}
Y.~Xu, W.~Xu, D.~Cheung, and Z.~Tu, ``Line segment detection using transformers
  without edges,'' in \emph{Proceedings of the IEEE/CVF Conference on Computer
  Vision and Pattern Recognition}, 2021, pp. 4257--4266.

\bibitem{kamat1998complete}
V.~Kamat-Sadekar and S.~Ganesan, ``Complete description of multiple line
  segments using the hough transform,'' \emph{Image and Vision Computing},
  vol.~16, no. 9-10, pp. 597--613, 1998.

\bibitem{wu2021deepcad}
R.~Wu, C.~Xiao, and C.~Zheng, ``Deepcad: A deep generative network for
  computer-aided design models,'' in \emph{Proceedings of the ICCV}, 2021, pp.
  6772--6782.

\bibitem{nash2020polygen}
C.~Nash, Y.~Ganin, S.~A. Eslami, and P.~Battaglia, ``Polygen: An autoregressive
  generative model of 3d meshes,'' in \emph{ICML}.\hskip 1em plus 0.5em minus
  0.4em\relax PMLR, 2020, pp. 7220--7229.

\bibitem{jones2020shapeassembly}
R.~K. Jones, T.~Barton, X.~Xu, K.~Wang, E.~Jiang, P.~Guerrero, N.~J. Mitra, and
  D.~Ritchie, ``Shapeassembly: Learning to generate programs for 3d shape
  structure synthesis,'' \emph{ACM Transactions on Graphics (TOG)}, vol.~39,
  no.~6, pp. 1--20, 2020.

\bibitem{groueix2018papier}
T.~Groueix, M.~Fisher, V.~G. Kim, B.~C. Russell, and M.~Aubry, ``A
  papier-m{\^a}ch{\'e} approach to learning 3d surface generation,'' in
  \emph{Proceedings of the CVPR}, 2018, pp. 216--224.

\bibitem{newell2016stacked}
A.~Newell, K.~Yang, and J.~Deng, ``Stacked hourglass networks for human pose
  estimation,'' in \emph{Proceedings of the ECCV}.\hskip 1em plus 0.5em minus
  0.4em\relax Springer, 2016, pp. 483--499.

\bibitem{xiao2018simple}
B.~Xiao, H.~Wu, and Y.~Wei, ``Simple baselines for human pose estimation and
  tracking,'' in \emph{Proceedings of the ECCV}, 2018, pp. 466--481.

\bibitem{zimmermann2017learning}
C.~Zimmermann and T.~Brox, ``Learning to estimate 3d hand pose from single rgb
  images,'' in \emph{Proceedings of the ICCV}, 2017, pp. 4903--4911.

\bibitem{lee2017roomnet}
C.-Y. Lee, V.~Badrinarayanan, T.~Malisiewicz, and A.~Rabinovich, ``Roomnet:
  End-to-end room layout estimation,'' in \emph{Proceedings of the ICCV}, 2017,
  pp. 4865--4874.

\bibitem{zou2018layoutnet}
C.~Zou, A.~Colburn, Q.~Shan, and D.~Hoiem, ``Layoutnet: Reconstructing the 3d
  room layout from a single rgb image,'' in \emph{Proceedings of the CVPR},
  2018, pp. 2051--2059.

\bibitem{liu2017raster}
C.~Liu, J.~Wu, P.~Kohli, and Y.~Furukawa, ``Raster-to-vector: Revisiting
  floorplan transformation,'' in \emph{Proceedings of the ICCV}, 2017, pp.
  2195--2203.

\bibitem{zeng2019deep}
Z.~Zeng, X.~Li, Y.~K. Yu, and C.-W. Fu, ``Deep floor plan recognition using a
  multi-task network with room-boundary-guided attention,'' in
  \emph{Proceedings of the ICCV}, 2019, pp. 9096--9104.

\bibitem{chen2022heat}
J.~Chen, Y.~Qian, and Y.~Furukawa, ``Heat: Holistic edge attention transformer
  for structured reconstruction,'' in \emph{Proceedings of CVPR}, 2022, pp.
  3866--3875.

\bibitem{zhang2021structured}
F.~Zhang, X.~Xu, N.~Nauata, and Y.~Furukawa, ``Structured outdoor architecture
  reconstruction by exploration and classification,'' in \emph{Proceedings of
  ICCV}, 2021, pp. 12\,427--12\,435.

\bibitem{li2021locally}
H.~Li, X.~Zheng, M.~Dong, G.-S. Xia, and H.~Xiong, ``Locally nonlinear affine
  verification for multisensor image matching,'' \emph{IEEE Transactions on
  Geoscience and Remote Sensing}, vol.~60, pp. 1--16, 2021.

\bibitem{zheng2022smoothly}
X.~Zheng, Z.~Yuan, Z.~Dong, M.~Dong, J.~Gong, and H.~Xiong, ``Smoothly varying
  projective transformation for line segment matching,'' \emph{ISPRS Journal of
  Photogrammetry and Remote Sensing}, vol. 183, pp. 129--146, 2022.

\bibitem{he2016deep}
K.~He, X.~Zhang, S.~Ren, and J.~Sun, ``Deep residual learning for image
  recognition,'' in \emph{Proceedings of the CVPR}, 2016, pp. 770--778.

\bibitem{nauata2020vectorizing}
N.~Nauata and Y.~Furukawa, ``Vectorizing world buildings: Planar graph
  reconstruction by primitive detection and relationship inference,'' in
  \emph{ECCV 2020}.\hskip 1em plus 0.5em minus 0.4em\relax Springer, 2020, pp.
  711--726.

\bibitem{van2018spacenet}
A.~Van~Etten, D.~Lindenbaum, and T.~M. Bacastow, ``Spacenet: A remote sensing
  dataset and challenge series,'' \emph{arXiv preprint arXiv:1807.01232}, 2018.

\bibitem{xue2020holistically}
N.~Xue, T.~Wu, S.~Bai, F.~Wang, G.-S. Xia, L.~Zhang, and P.~H. Torr,
  ``Holistically-attracted wireframe parsing,'' in \emph{Proceedings of the
  IEEE/CVF Conference on Computer Vision and Pattern Recognition}, 2020, pp.
  2788--2797.

\bibitem{Yang2016Object}
J.~Yang, B.~Price, S.~Cohen, H.~Lee, and M.-H. Yang, ``Object contour detection
  with a fully convolutional encoder-decoder network,'' in \emph{CVPR}, 2016,
  pp. 193--202.

\bibitem{stekovic2021montefloor}
S.~Stekovic, M.~Rad, F.~Fraundorfer, and V.~Lepetit, ``Montefloor: Extending
  mcts for reconstructing accurate large-scale floor plans,'' in
  \emph{Proceedings of the IEEE/CVF International Conference on Computer
  Vision}, 2021, pp. 16\,034--16\,043.

\bibitem{lv2021residential}
X.~Lv, S.~Zhao, X.~Yu, and B.~Zhao, ``Residential floor plan recognition and
  reconstruction,'' in \emph{Proceedings of the IEEE/CVF Conference on Computer
  Vision and Pattern Recognition}, 2021, pp. 16\,717--16\,726.

\bibitem{lcnn}
Y.~Zhou, H.~Qi, and Y.~Ma, ``End-to-end wireframe parsing,'' in \emph{2019
  IEEE/CVF International Conference on Computer Vision (ICCV)}, 2019, pp.
  962--971.

\end{thebibliography}
}

\end{document}